\documentclass[letterpaper,10pt,conference]{ieeeconf}
\IEEEoverridecommandlockouts
\usepackage{times}
\usepackage{flushend}
\usepackage{amsmath}
\usepackage{amssymb}
\usepackage{graphicx}
\usepackage{booktabs}
\usepackage{multirow}
\usepackage{array}
\usepackage{tabularx}
\usepackage[table]{xcolor}
\usepackage{float}
\usepackage{placeins}
\usepackage{cite}
\usepackage[bookmarks=true]{hyperref}

\hypersetup{
    pdftitle={CoPRE: Improving Sensitivity in Proprioceptive Contact Detection for Low-Cost Robot Arms},
    pdfauthor={Author Names Omitted for Anonymous Review},
    pdfsubject={Proprioceptive weak-contact detection for contact-aware robot action selection},
    pdfkeywords={proprioceptive contact sensing, weak contact, contact-aware action selection, manipulation safety}
}

\title{\LARGE \bf
CoPRE: Improving Sensitivity in Proprioceptive Contact Detection for Low-Cost Robot Arms}
\author{Author Names Omitted for Anonymous Review\\[5.865pt]\mbox{}}

\author{
\authorblockN{
Yuxiao Zhu\textsuperscript{1,*}\quad
Jinzhou Li\textsuperscript{1,*}\quad
Yifei Dong\textsuperscript{2}\quad
Muhammad Suhail\textsuperscript{3}\quad
Chunyuan Yang\textsuperscript{1}\\[0.35em]
Xinyuan Luo\textsuperscript{1}\quad
Haoyu Li\textsuperscript{1}\quad
Boyuan Chen\textsuperscript{1}\quad
Xianyi Cheng\textsuperscript{1}
}
\vspace{0.5em}
\authorblockA{
\textsuperscript{1}Duke University
\qquad
\textsuperscript{2}KTH Royal Institute of Technology
\qquad
\textsuperscript{3}Carnegie Mellon University
\qquad
\textsuperscript{*}Equal contribution
}}

\definecolor{oursblue}{RGB}{235,244,252}

\begin{document}
\bstctlcite{IEEEcontrol}
\maketitle
\thispagestyle{empty}
\pagestyle{empty}

\begin{abstract}

Contact detection during robotic manipulation allows robots to recognize unexpected contact and adapt their motion accordingly. However, in low-cost robot arms without dedicated force or tactile sensors, detecting weak contacts from proprioception is challenging because the resulting changes in joint-level proprioceptive signals can be small compared to normal variation and noise caused by robot motion itself. We introduce Contact-free Proprioceptive Response Estimation (CoPRE), improving proprioceptive contact detection sensitivity using only contact-free motion, without additional force sensors, contact labels, or analytical dynamics models. 
CoPRE estimate the expected joint torques under contact-free motion from proprioceptive state history and commanded motion, while removing recent observations that may already reflect contact. 
It then computes the residual between the expected and observed joint torque estimates, and maps this residual to a contact score using a noise-weighted Jacobian. 
Real-robot experiments on ARX Arm and Unitree G1 show that CoPRE achieves 74.1\% and 82.2\% recall on the tested contact trials, compared with 0\%/0\% on ARX and 16.3\%/42.2\% on G1 for the learned torque-prediction and inverse-dynamics baselines. CoPRE also reaches 90\% detection rate for pushing force at 3.5 N on ARX and 5.5 N on G1.
To demonstrate the downstream utility of our method, we implement belief-space manipulation planning for obstacle-aware object placement and book insertion where detected contacts update the spatial belief and enable the robot to retreat from blocked motions, adjust its pose, and retry. Project website: \url{https://copre-arm.github.io}.

\end{abstract}

\IEEEpeerreviewmaketitle

\section{Introduction}
\label{sec:introduction}

\begin{figure}[!t]
    \centering
    \includegraphics[width=\columnwidth,trim=0 8bp 7.5bp 0,clip]{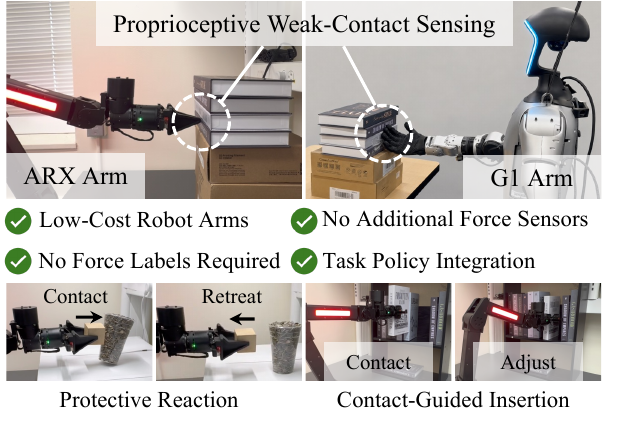}
    \caption{\textbf{Weak-contact detection and reactive manipulation with CoPRE.} Built-in joint states and torque estimates provide contact feedback on ARX and G1, using contact-free training data without additional force sensors or force labels. Bottom: on ARX, detected contact triggers protective retreat during placement (left) and pose adjustment during book insertion (right).}
    \label{fig:teaser}
    \vspace{-0.3cm}
\end{figure}

\looseness=-1 Contact detection helps robots recognize unexpected interactions and adapt their motion during manipulation~\cite{suomalainen2022survey,bohg2017interactive}. Uncertainty in object poses and local geometry can cause planned placement or insertion motions to encounter unexpected obstacles. Contact feedback lets the robot stop, retreat, and plan another attempt (Fig.~\ref{fig:teaser}). Low-cost robot arms can lack dedicated force or tactile sensors~\cite{zhao2023lowcost}; built-in joint states and motor signals offer contact feedback~\cite{deluca2005sensorless,dallalibera2019proprioceptive}.

\looseness=-1 Weak contact is difficult to distinguish from normal motion and noise. We define \emph{weak contact} by contact-induced joint torque changes that are small relative to contact-free variability. Residual-based detectors compare observed torque or motor current with a nominal reference to obtain a residual. The reference comes from dynamics or learned contact-free motion~\cite{deluca2005sensorless,dallalibera2019proprioceptive,oh2026factr2}. When contact-affected states enter the predictor, the learned reference may follow the torque change and reduce the residual. Prediction errors also produce contact-free residuals that vary across joints and motions. Suppressing false alarms from these residuals can require higher thresholds that miss weak contacts. Detection must preserve contact-induced deviations and account for contact-free residual variability.

\looseness=-1 We introduce Contact-free Proprioceptive Response Estimation (CoPRE) to address these difficulties. CoPRE predicts the joint torque expected during contact-free motion from earlier states and commanded motion. State exclusion prevents observations within the prediction interval from entering the predictor, preserving the reference against recent contact-induced state changes. The difference between observed and predicted torque is mapped to a contact score through noise-weighted Jacobian aggregation. The aggregation gives less weight to joints with larger contact-free residual variability. Training and calibration use only contact-free motion, without additional force sensors, contact labels, or analytical dynamics models.

\looseness=-1 We integrate CoPRE with belief-guided planning for object placement and book insertion on ARX. When contact blocks a motion, the robot stops and updates a spatial belief map using the contact decision, robot pose, and task progress. Contact provides evidence of a nearby obstruction, while traversed paths provide evidence of free space. After retreat, the map guides the next placement or insertion attempt.

\looseness=-1 Book-pushing experiments compare CoPRE with neural and dynamics baselines on ARX L5 and the Unitree G1 right arm. Combined calibration uses task-matched and broad contact-free motion under shared rules within each robot. ARX recall is 74.1\% for CoPRE and zero for both baselines. On G1, CoPRE achieves 82.2\%, compared with 16.3\% for the neural and 42.2\% for the dynamics baseline. Detection reaches at least 90\% at separately measured reference sliding resistances of 3.5~N and 5.5~N, respectively, and at all higher tested levels. Matched ablations support the roles of state exclusion and noise weighting. The ARX manipulation demonstrations provide qualitative evidence of downstream use.

Our contributions are twofold:
\begin{itemize}
    \item \textbf{Improved sensitivity in proprioceptive contact detection.}
    \looseness=-1 CoPRE uses state exclusion to preserve contact-induced deviations and noise-weighted Jacobian aggregation to account for residual variability.
    \item \textbf{Real-robot implementation, validation, and downstream integration.}
    \looseness=-1 We implement and evaluate CoPRE on ARX L5 and G1 through baseline comparisons and matched ablations, and integrate its contact feedback with belief-guided manipulation on ARX.
\end{itemize}

\section{Related Work}
\label{sec:related_work}

\subsection{Model-based contact and force estimation}
Model-based methods detect contact and estimate external forces from deviations from nominal dynamics~\cite{deluca2005sensorless,deluca2006collision,haddadin2017collisions,wahrburg2018motor}. Model errors, friction, and noise limit sensitivity. Hybrid approaches use semiparametric dynamics learning and disturbance filtering~\cite{hu2018semiparametric}, or recurrent learning of momentum-observer uncertainty~\cite{lim2021momentum}. CoPRE instead learns nominal joint torque from contact-free recordings, reducing reliance on identified dynamics and explicit friction models while retaining a kinematic model for residual aggregation.

\subsection{Learning-based contact and force sensing}
Learning-based methods model nominal behavior or learn contact outputs from labeled interactions. Learned current and inverse-dynamics models support residual-based sensing~\cite{dallalibera2019proprioceptive,yilmaz2020inverse,shan2024sensorless}; unsupervised anomaly detection identifies departures from normal motion~\cite{park2022unsupervised}. NEXT in FACTR 2 predicts free-space torque from state and tracking-error histories extending to the current time~\cite{oh2026factr2}. Supervised methods learn collision detection~\cite{park2021collision,kim2022transferable,niu2024madcnn}, localization~\cite{liang2021contact,fu2025unitac}, or force prediction~\cite{dou2026neuralactuator}. CoPRE uses no contact or force labels, excludes prediction-interval states, and combines noise-weighted aggregation with separate contact-free calibration.

\subsection{Contact sensing and feedback for manipulation}
Contact feedback supports information gathering and motion adaptation~\cite{bohg2017interactive,suomalainen2022survey}. Tactile feedback guides motion through clutter and helps resolve jamming~\cite{jain2013manipulation,brouwer2024tactileprimitives}. Contact observations also support object tracking, active search, and blind retrieval~\cite{zhong2022stucco,zhong2025rumi,saleem2025contactdriven}, while haptic planning has been applied to crowded bookshelf insertion~\cite{yang2025hapticmetric}. Feedback controllers regulate interaction using force measurements or actuator signals~\cite{jordana2024ffmpc,shi2026minimalist,ma2026current}; Blind Dexterity learns whole-body manipulation from proprioceptive histories and purposeful contact~\cite{bhatt2026blind}. We evaluate CoPRE on ARX and G1, with belief-guided placement and insertion on ARX.

\begin{figure*}[!t]
    \centering
    \includegraphics[width=\textwidth,trim=0 0.25cm 0 0,clip]{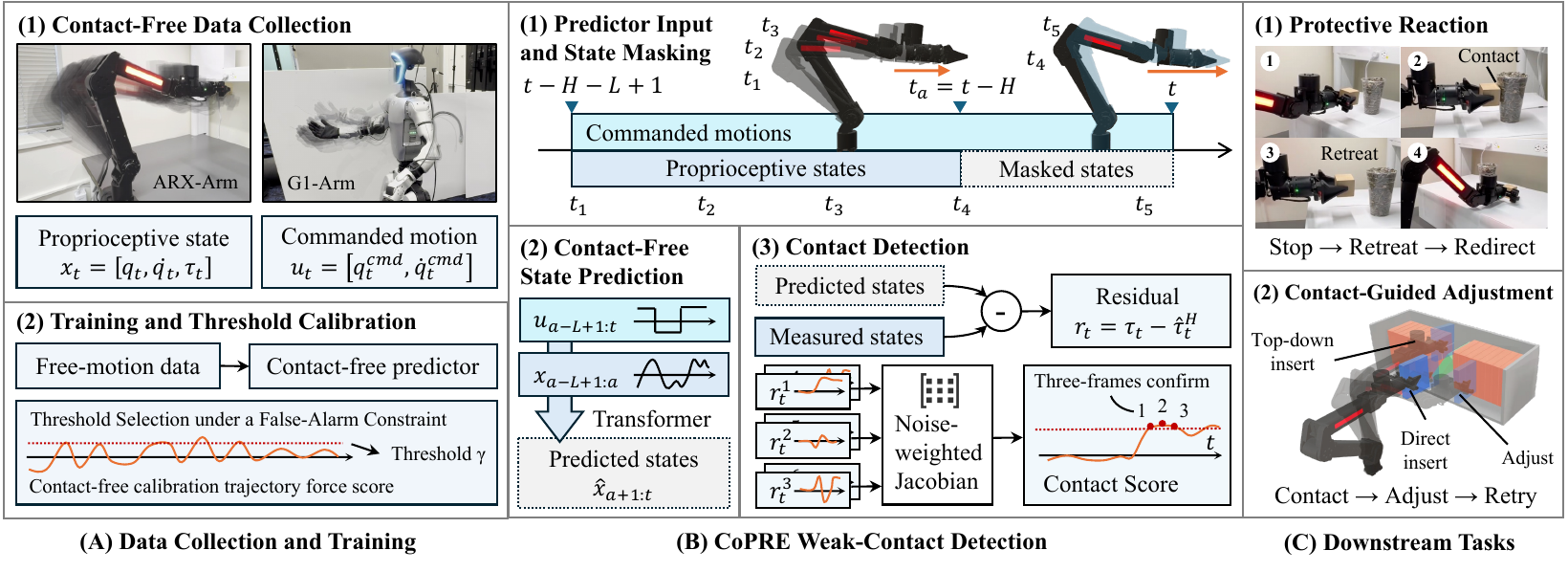}
    \caption{\textbf{CoPRE: from proprioception to contact-guided action.} (A) Contact-free recordings train the nominal-response predictor and calibrate the detection threshold. (B) A history of $L$ state--command pairs and known commands predicts $H$ states, excluding observations within that interval. Torque residuals yield a contact score through bias correction and noise-weighted Jacobian aggregation. The schematic omits bias correction. Three consecutive samples at or above threshold confirm contact. (C) Contact feedback guides retreat or pose adjustment and retry.}
    \label{fig:method}
    \vspace{-0.4cm}
\end{figure*}

\section{Problem Definition}
\label{sec:problem}

\looseness=-1 We study contact detection from built-in proprioceptive signals. We focus on weak contact as defined above, seeking to distinguish contact-induced torque changes from contact-free variability. Our experiments cover contact at the end effector or through a grasped object. For a $d$-joint robot, $x_t$ and $u_t$ denote the observed state and joint commands at time $t$:
\begin{equation}
    x_t=[q_t,\dot q_t,\tau_t] \in \mathbb{R}^{3d}, \qquad
    u_t=[q_t^{\mathrm{cmd}},\dot q_t^{\mathrm{cmd}}] \in \mathbb{R}^{2d}.
    \label{eq:signals}
\end{equation}
Here, $q_t$ and $\dot q_t$ are joint position and velocity. The observed torque $\tau_t$ is the controller's joint torque estimate, obtained from built-in motor signals on our platforms. $q_t^{\mathrm{cmd}}$ and $\dot q_t^{\mathrm{cmd}}$ are the commanded joint position and velocity.

\looseness=-1 Let $c_t\in\{0,1\}$ indicate whether external contact is present at step $t$; this quantity is unavailable to the detector. A causal detector maps these states and commands to a nonnegative score $s_t$ and binary contact decision $\hat c_t$:
\begin{equation}
    s_t=g_\theta(x_{1:t},u_{1:t}), \qquad
    \hat c_t=\mathcal{D}_\gamma(s_{1:t})\in\{0,1\},
    \label{eq:detection_problem}
\end{equation}
where $g_\theta$ maps proprioceptive observations and joint commands to a contact score $s_t$, and $\mathcal{D}_\gamma$ converts the score history into a binary contact decision by requiring consecutive samples at or above threshold $\gamma$. A false alarm occurs when $\hat c_t=1$ during contact-free motion ($c_t=0$). The goal is high weak-contact sensitivity while limiting false alarms.

\section{CoPRE}
\label{sec:method}

CoPRE converts joint observations and commands into contact feedback for manipulation (Fig.~\ref{fig:method}). A model trained on contact-free recordings predicts nominal joint torque: the controller's expected torque estimate during contact-free motion. We subtract this prediction from the observed torque estimate and aggregate the bias-corrected residuals through a noise-weighted Jacobian to obtain a contact score. The rule $\mathcal{D}_\gamma$ converts this score history into the contact decision $\hat c_t$.

\subsection{Contact-Free Data, Training, and Calibration}

Contact-free recordings establish the nominal motion reference. We record states and commands (Eq.~\eqref{eq:signals}, Fig.~\ref{fig:method}(A1)) for each robot. Separate training, validation, and calibration sets fit its predictor, select the checkpoint with the lowest validation loss, and set the detection threshold.

\looseness=-1 For a prediction horizon of $H$ steps, the predictor learns the next $H$ states from earlier observations and joint commands (Fig.~\ref{fig:method}(A2)). Inputs and targets are normalized using training statistics. Let $\ell_q$, $\ell_{\dot q}$, and $\ell_\tau$ denote mean-squared errors over all predicted steps and joints in normalized coordinates. The loss is
\begin{equation}
    \mathcal{L}=\ell_\tau+\alpha_q\ell_q+\alpha_{\dot q}\ell_{\dot q},
    \label{eq:training_loss}
\end{equation}
Joint position and velocity predictions provide auxiliary supervision; detection uses only the final-step torque prediction. The experiments specify the loss weights. Training residuals set the bias and scale for detection.

\looseness=-1 Even without contact, prediction errors produce nonzero scores. We therefore calibrate the detector on separate contact-free recordings, choosing the lowest threshold that satisfies the prescribed false-alarm constraint under the same confirmation rule used at deployment. The experiments specify each robot's constraint. The model, statistics, and threshold remain fixed; new motions may alter false-alarm rates.

\subsection{Nominal State Prediction}

Contact can change the measured state. If that state enters the predictor, the predicted nominal joint torque may also change, leaving a smaller residual for detection. CoPRE uses state exclusion to predict from an earlier state history (Fig.~\ref{fig:method}(B1)). It predicts $H$ states from $L$ samples ending at $t_a=t-H$:
\begin{equation}
    \hat x_{t_a+1:t}
    = f_\theta(x_{t_a-L+1:t_a},u_{t_a-L+1:t_a},u_{t_a+1:t}).
    \label{eq:copre_contract}
\end{equation}
The anchor $t_a$ is the last step whose state enters the predictor. State exclusion withholds observations over $(t_a,t]$ while retaining commands. Detection uses current torque and joint positions to compute the residual and Jacobian. Contact can still affect earlier states and commands.

To implement Eq.~\eqref{eq:copre_contract}, a transformer~\cite{vaswani2017attention} encodes $L$ state--command pairs with learned positional embeddings to mark their order (Fig.~\ref{fig:method}(B2)). The decoder predicts $H$ states using queries that share the anchor state, each paired with its own command and learned step embedding. A causal mask lets each query attend only to its own and earlier steps. The decoder adds a predicted change to the anchor state without feeding predicted states back as inputs. As the window advances, detection uses only the predicted nominal joint torque at the final step, $\hat\tau_t^{(H)}$.

\begin{figure*}[!t]
    \centering
    \includegraphics[width=\textwidth]{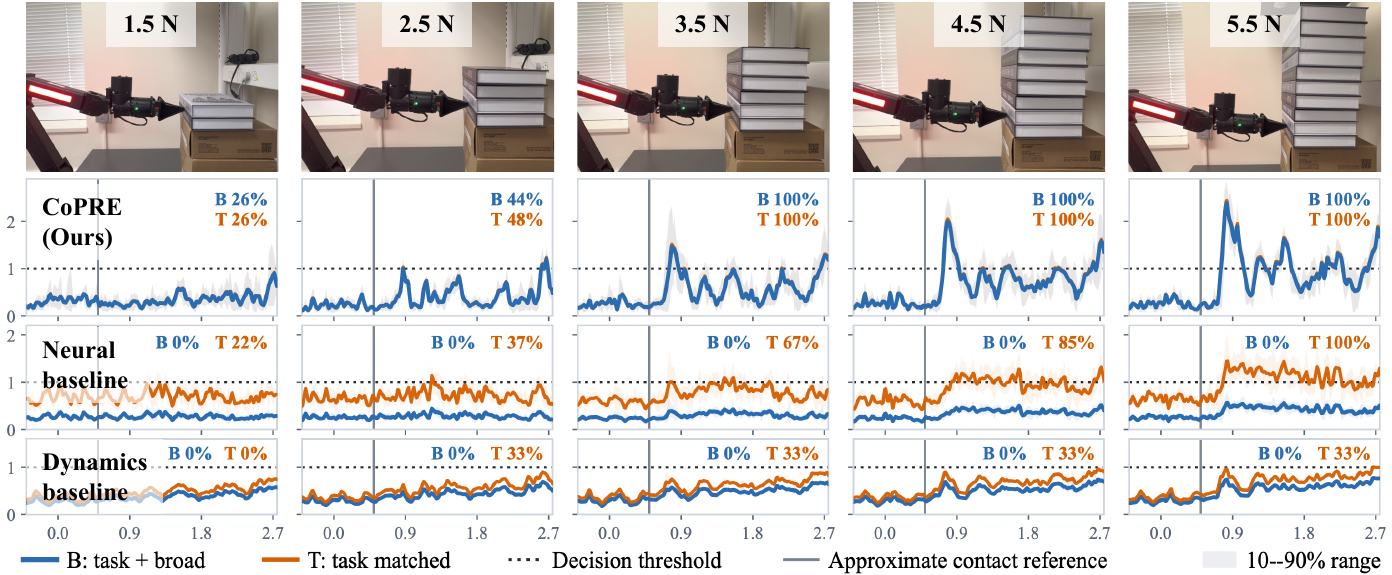}
    \caption{\textbf{Weak-contact detection on ARX.} Columns show separately measured reference sliding resistances of the book stacks. Curves show threshold-normalized scores at 0.04~m/s: mean and 10--90\% range over trials and model seeds. Dotted lines mark threshold 1; vertical lines mark geometric contact references. Percentages report trial-level detection across all three pushing speeds. \textcolor[RGB]{213,94,0}{Orange (T)} uses task-matched contact-free motion calibration; \textcolor[RGB]{43,108,176}{blue (B)} adds broad contact-free motion. Thresholds permit zero confirmed calibration alarms. Only CoPRE retains detections under combined calibration.}
    \label{fig:calibration_force_levels}
\end{figure*}

\subsection{Noise-Weighted Contact Detection}

Joints with larger residual variability during contact-free motion provide less reliable evidence of contact. We reduce their fitting weights and use the Jacobian to combine joint residuals into a contact score (Fig.~\ref{fig:method}(B3)). Let $r_t$ be the difference between observed and predicted joint torques after correcting the nominal bias $b$. The translational end-effector Jacobian $J_t\in\mathbb{R}^{3\times d}$, evaluated at $q_t$, maps a force vector to joint torques through $J_t^\top$. We fit the residual by weighted damped least squares:
\begin{equation}
\begin{aligned}
    r_t &= \tau_t-\hat{\tau}_t^{(H)}-b, \qquad s_t=\lVert\hat f_t\rVert_2, \\
    \hat f_t &= \arg\min_f
    \left\lVert D_\sigma^{-1}(J_t^\top f-r_t)\right\rVert_2^2
    +\lambda_t\lVert f\rVert_2^2,
\end{aligned}
    \label{eq:force_proxy}
\end{equation}
\looseness=-1 The fitted vector $\hat f_t\in\mathbb{R}^3$ is an equivalent end-effector force, not a validated contact-force measurement. Both $\hat f_t$ and $J_t$ use the kinematic model's world frame. Predictions are restored to the controller's torque units; $b$ is the mean training residual at step $H$. The diagonal $D_\sigma$ contains robust residual scales $\sigma_j=\max(1.4826\,\mathrm{MAD}_j,10^{-4})$, where MAD is the median absolute deviation. The positive scales reflect prediction error, motion effects, and noise.

Damping limits large solutions near poorly conditioned Jacobians and scales with the weighted Jacobian:
\begin{equation*}
    \lambda_t=\max\!\left(\epsilon,\lambda\max\!\left[\frac{\mathrm{tr}(J_{t}D_\sigma^{-2}J_t^\top)}{3},\epsilon\right]\right),
\end{equation*}
\looseness=-1 where $\lambda$ sets relative damping and $\epsilon$ is a numerical floor. The ablations call $s_t$ the weighted Jacobian score. This scalar detects contact; force accuracy and localization remain untested.

To reject isolated score excursions, $K$ consecutive samples at or above $\gamma$ set the binary contact decision to 1 (Fig.~\ref{fig:method}(B3)). A lower score resets the decision and counter. We normalize the score as $s_t/\gamma$, with threshold 1.

\section{Experiments}
\label{sec:experiments}
\looseness=-1 We evaluate CoPRE through three questions on contact detection, design choices, and downstream manipulation:
\begin{enumerate}
    \item[\textbf{Q1}] {\raggedright How sensitive is CoPRE to weak contact compared \mbox{with existing methods?}\par}
    \item[\textbf{Q2}] How do CoPRE’s design choices affect contact detection sensitivity and false alarms?
    \item[\textbf{Q3}]How can sensitive proprioceptive contact detection guide manipulation without additional force sensing?
\end{enumerate}

\subsection{Detection Benchmark Setup}

Q1 and Q2 use the detection benchmark on both arms; Q3 evaluates separate manipulation tasks.

\begin{figure*}[!t]
\centering
\includegraphics[width=\textwidth]{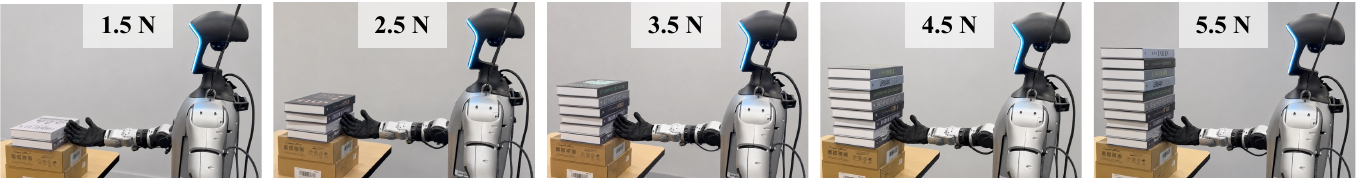}\par
\vspace{-0.25ex}

\begin{minipage}[t]{0.495\textwidth}
\centering
\footnotesize
\setlength{\tabcolsep}{1.5pt}
\renewcommand{\arraystretch}{1.04}
\textbf{(a) Detection rate (\%) by reference sliding resistance (N)}\par
\vspace{0.35ex}
\begin{tabularx}{\linewidth}{@{}>{\raggedright\arraybackslash}p{0.29\linewidth}*{5}{>{\centering\arraybackslash}X}@{}}
\toprule
Method & 1.5 & 2.5 & 3.5 & 4.5 & 5.5 \\
\midrule
Dynamics baseline & 22.2 & 22.2 & 55.6 & 44.4 & 66.7 \\
Neural baseline & 0.0 & 18.5 & 0.0 & 25.9 & 37.0 \\
\rowcolor{oursblue}
\textbf{CoPRE (Ours)} & \textbf{55.6} & \textbf{85.2} & \textbf{85.2} & \textbf{85.2} & \textbf{100.0} \\
\bottomrule
\end{tabularx}
\end{minipage}%
\hfill
\begin{minipage}[t]{0.495\textwidth}
\centering
\footnotesize
\setlength{\tabcolsep}{1.5pt}
\renewcommand{\arraystretch}{1.04}
\textbf{(b) Operating characteristics}\par
\vspace{0.35ex}
\begin{tabularx}{\linewidth}{@{}>{\raggedright\arraybackslash}p{0.265\linewidth}*{4}{>{\centering\arraybackslash}X}@{}}
\toprule
Method & {\scriptsize $F_{90}$ (N)} & \makebox[\linewidth][c]{\scriptsize Recall (\%)} &
{\scriptsize Task FA$_t$} & {\scriptsize Broad FA$_t$} \\
\midrule
Dynamics baseline & $>5.5$ & 42.2 & 0.80 & 0.36 \\
Neural baseline & $>5.5$ & 16.3 & 0.37 & 1.51 \\
\rowcolor{oursblue}
\textbf{CoPRE (Ours)} & $\mathbf{5.5}$ & \textbf{82.2} & 1.37 & 0.86 \\
\bottomrule
\end{tabularx}
\end{minipage}

\caption{\textbf{Weak-contact detection on the G1 right arm.} Top: book stacks with separately measured reference sliding resistances. (a) Trial-level detection within 0.5~s of the geometric contact reference, pooled across three speeds and averaged over model seeds. (b) $F_{90}$ is the lowest tested resistance with at least 90\% detection at that and all higher levels; $>5.5$ means it was not reached. Recall pools all trials. Task/Broad FA$_t$ is held-out task-matched/broader contact-free time in confirmed alarm (\%). All methods share a 1\% calibration FA$_t$ budget. CoPRE reaches $F_{90}=5.5$~N.}
\label{fig:g1_results}
\end{figure*}

\vspace{1mm}
\noindent\textbf{Robot platforms.}
We evaluate two arm platforms: a six-degree-of-freedom ARX L5 manipulator~\cite{arx_robotics_website} and the seven-joint right arm of the Unitree G1 humanoid~\cite{unitree_g1_website} (Figs.~\ref{fig:calibration_force_levels} and~\ref{fig:g1_results}). During pushing, G1 stands autonomously under the SONIC policy~\cite{luo2025sonic}. Detection uses only right-arm signals; leg, waist, left-arm, and floating-base measurements are excluded. We read joint position, velocity, and torque estimates.

\looseness=-1 Detection uses only built-in proprioception. On ARX, the SDK estimates joint torque by scaling motor-current feedback with motor-specific factors. On G1, we read \texttt{tau\_est} from the low-level motor-state interface. Neither arm uses dedicated torque sensors. Current-derived estimates are subject to scaling error, temperature, friction, and transmission effects~\cite{oh2026factr2}.

\noindent\textbf{Detection task.}
\looseness=-1 Each arm pushes five book stacks at $0.02$, $0.04$, and $0.06$~m/s, with three repetitions per condition: nine physical trials per reference sliding resistance and 45 per arm. The stacks weigh $300$, $610$, $930$, $1270$, and $1610$~g. Separate measurements of the approximate peak force needed to initiate sliding give reference sliding resistances $F_{\mathrm{ref}}$ of $1.5$, $2.5$, $3.5$, $4.5$, and $5.5$~N. These separately measured values characterize each pushing condition but do not track force during individual trials. We collect task-matched contact-free motion by repeating the pushing trajectory without an object. Broad contact-free motion varies pose, direction, and speed, with stops and reversals.

\vspace{1mm}
\noindent\textbf{Data separation.}
\looseness=-1 Training recordings fit the predictors and residual statistics. Validation recordings select the saved models with the lowest validation loss; separate calibration recordings set the detection thresholds. Contact trials are excluded from fitting and calibration. Within each platform, all methods use the same recordings and evaluation splits. Learned models use three seeds; dynamics is evaluated once. The default horizon $H=3$ was selected on data separate from the reported contact trials.

\vspace{1mm}
\noindent\textbf{Metrics.}
\looseness=-1 We report trial-level detection at each reference sliding resistance. Only alarms confirmed within the detection evaluation window count: from the geometric contact reference time $t_{\mathrm{ref}}$ through push end on ARX, or for 0.5~s on G1. Confirmation restarts at $t_{\mathrm{ref}}$, the recorded contact-plane crossing, not independently measured physical contact onset. Methods share the evaluation window within each platform. Detection rates average trial outcomes across seeds; overall trial recall $R$ pools all resistances and speeds. Seeds do not add physical trials. We define $F_{90}$ as the lowest reference sliding resistance with at least 90\% detection at that and all higher tested levels. If none qualifies, we report $F_{90}>5.5$~N. This metric characterizes sensitivity under the calibration rule; force at detection is unmeasured. Recall uses 95\% bootstrap confidence intervals over physical trials, grouping their seed evaluations. The false alarm time fraction FA$_t$ (\%) measures how much of the monitored contact-free time is spent in confirmed alarm.

\vspace{1mm}
\noindent\textbf{Baselines.}
\looseness=-1 We compare CoPRE with a neural torque predictor based on NEXT~\cite{oh2026factr2} and a nominal inverse-dynamics detector. The neural baseline uses a two-layer, 128-unit LSTM over joint position, velocity, and position tracking error. The ARX implementation uses 50 history samples and torque MSE; the G1 implementation uses 25 history samples and Smooth L1 loss. Both use the L1 norm of the unweighted torque residual as the contact score. We evaluate NEXT adapted to each arm under the shared protocol. The dynamics baseline computes nominal joint torques from inertia, gravity, Coriolis and centrifugal effects, and available joint-friction parameters. It subtracts these torques from the joint torque estimates and fits the residual using unweighted damped least squares with the translational Jacobian. Existing parameters are used without additional dynamics identification. To estimate acceleration, we causally filter velocity with a 0.10~s time constant and take the backward difference. ARX has no active joint-friction parameters; G1 uses nominal friction and damping with its base and non-arm joints held fixed in the dynamics model.

\vspace{1mm}
\noindent\textbf{Calibration protocols.}
All methods confirm contact after three consecutive samples at or above the threshold. Thresholds are selected independently under a common rule on each platform. On ARX, we select the lowest threshold with zero confirmed false alarms on the calibration set. Task-matched calibration uses nine contact-free executions of the pushing trajectory; combined calibration uses the union of these recordings and one trajectory of diverse contact-free motions. We keep the models and contact trials fixed to isolate the effect of calibration coverage.

\looseness=-1 G1 uses three-fold cross-fitting over contact-free motion recordings. Each fold calibrates on nine task-matched and three broad contact-free runs, while holding out three task-matched runs and one broad contact-free run for FA evaluation. The threshold is the lowest value with at most 1\% FA$_t$ in each calibration motion set. The corresponding contact trials are used only for evaluation within that fold, and all methods share the splits. We selected this budget during development from a sweep of 1\%, 2\%, 3\%, 5\%, and 10\%. Held-out false alarms are reported separately on new executions of the task and broad contact-free trajectories. The different ARX and G1 rules restrict comparisons to each platform.

\vspace{1mm}
\noindent\textbf{Implementation Details.}
\looseness=-1 Both arms use $L=20$, $H=3$, $K=3$, $\alpha_q=\alpha_{\dot q}=0.1$, and $\epsilon=10^{-12}$, with a 96-dimensional transformer with four attention heads, two encoder and two decoder layers, and feedforward width 192. We train with AdamW~\cite{loshchilov2019adamw} at learning rate $5\times10^{-4}$, weight decay $10^{-4}$, and batch size 256 for up to 60 epochs, stopping after 10 epochs without validation improvement. Seeds are 7, 17, and 27, and relative Jacobian damping is $\lambda=10^{-3}$. Windows use consecutive samples at typical intervals of 32\,ms on ARX and 20\,ms on G1. ARX training uses 18 task-matched contact-free and two broad contact-free runs. Each G1 fold uses 24 task-matched contact-free and two broad contact-free runs for training and nine separate task-matched contact-free runs for checkpoint selection. The six G1 broad contact-free recordings last 120~s each: two for training and four for cross-fitted calibration and evaluation. Calibration excludes the first 0.5\,s.

\begin{figure*}[!t]
\centering
\scriptsize
\renewcommand{\arraystretch}{1.04}
\textbf{(a) State exclusion and residual aggregation}\par
\vspace{0.2ex}
\setlength{\tabcolsep}{1.1pt}
\begin{tabularx}{\textwidth}{@{}ll>{\hsize=0.95\hsize\linewidth=\hsize\centering\arraybackslash}X>{\hsize=0.95\hsize\linewidth=\hsize\centering\arraybackslash}X>{\hsize=1.1\hsize\linewidth=\hsize\centering\arraybackslash}X>{\hsize=1\hsize\linewidth=\hsize\centering\arraybackslash}X>{\hsize=0.95\hsize\linewidth=\hsize\centering\arraybackslash}X>{\hsize=0.95\hsize\linewidth=\hsize\centering\arraybackslash}X>{\hsize=1.1\hsize\linewidth=\hsize\centering\arraybackslash}X>{\hsize=0.95\hsize\linewidth=\hsize\centering\arraybackslash}X>{\hsize=0.95\hsize\linewidth=\hsize\centering\arraybackslash}X>{\hsize=1.1\hsize\linewidth=\hsize\centering\arraybackslash}X>{\hsize=1\hsize\linewidth=\hsize\centering\arraybackslash}X>{\hsize=0.95\hsize\linewidth=\hsize\centering\arraybackslash}X>{\hsize=0.95\hsize\linewidth=\hsize\centering\arraybackslash}X>{\hsize=1.1\hsize\linewidth=\hsize\centering\arraybackslash}X@{}}
\toprule
\multirow{2}{*}{\shortstack[l]{State\\exclusion}} & \multirow{2}{*}{Aggregation} &
\multicolumn{3}{c}{ARX: Task-matched ($\gamma_T$)} & \multicolumn{4}{c}{ARX: Combined ($\gamma_B$)} &
\multicolumn{3}{c}{G1: Task-matched ($\gamma_T$)} & \multicolumn{4}{c}{G1: Combined ($\gamma_B$)} \\
\cmidrule(lr){3-5}\cmidrule(lr){6-9}\cmidrule(lr){10-12}\cmidrule(lr){13-16}
& & $F_{90}$ & R & \mbox{Broad FA$_t$} & $\gamma_{\rm br}/\gamma_T$ & $F_{90}$ & R & \mbox{Broad FA$_t$} & $F_{90}$ & R & \mbox{Broad FA$_t$} & $\gamma_{\rm br}/\gamma_T$ & $F_{90}$ & R & \mbox{Broad FA$_t$} \\
\midrule
Off & W. Jac. & $>5.5$ & $40.0$ & $4.67$ & $1.52$ & $>5.5$ & $3.7$ & $0.42$ & $>5.5$ & $65.9$ & $0.88$ & $0.96$ & $>5.5$ & $64.4$ & $0.84$ \\
Off & Std. norm & $>5.5$ & $56.3$ & $74.61$ & $3.53$ & $>5.5$ & $0.0$ & $0.63$ & $>5.5$ & $12.6$ & $1.82$ & $1.06$ & $>5.5$ & $9.6$ & $1.11$ \\
On & Std. norm & $3.5$ & $88.9$ & $85.48$ & $2.60$ & $>5.5$ & $0.0$ & $11.79$ & $>5.5$ & $49.6$ & $5.40$ & $1.18$ & $>5.5$ & $31.9$ & $1.02$ \\
\rowcolor{oursblue}
\textbf{On} & \textbf{W. Jac.} & $3.5$ & $74.8$ & $3.68$ & $1.02$ & $3.5$ & $74.1$ & $3.32$ & $5.5$ & $83.0$ & $0.92$ & $0.94$ & $5.5$ & $82.2$ & $0.86$ \\
\bottomrule
\end{tabularx}
\par
\vspace{0.75ex}

\begin{minipage}[t]{0.329\textwidth}
\centering
\textbf{(b) Aggregation and noise weighting}\par
\vspace{0.2ex}
\setlength{\tabcolsep}{1.1pt}
\begin{tabularx}{\linewidth}{@{}l*{6}{>{\raggedleft\arraybackslash}X}@{}}
\toprule
& \multicolumn{3}{c}{ARX} & \multicolumn{3}{c}{G1} \\
\cmidrule(lr){2-4}\cmidrule(lr){5-7}
Score & $F_{90}$ & R & FA$_t$ & $F_{90}$ & R & FA$_t$ \\
\midrule
Std. norm & $>5.5$ & $0.0$ & $11.79$ & $>5.5$ & $31.9$ & $1.02$ \\
U. Jac. & $5.5$ & $45.9$ & $1.67$ & $>5.5$ & $12.6$ & $0.38$ \\
\rowcolor{oursblue}
\textbf{W. Jac.} & $3.5$ & $74.1$ & $3.32$ & $5.5$ & $82.2$ & $0.86$ \\
\bottomrule
\end{tabularx}
\end{minipage}%
\hspace{0.012\textwidth}%
\begin{minipage}[t]{\dimexpr\textwidth-0.309\textwidth-0.329\textwidth-2\dimexpr0.012\textwidth\relax\relax}
\centering
\textbf{(c) G1 state exclusion at $H=2$--$4$}\par
\vspace{0.2ex}
\setlength{\tabcolsep}{1.1pt}
\begin{tabularx}{\linewidth}{@{}l*{6}{>{\raggedleft\arraybackslash}X}@{}}
\toprule
& \multicolumn{3}{c}{Off} & \multicolumn{3}{c}{On} \\
\cmidrule(lr){2-4}\cmidrule(lr){5-7}
Variant & $F_{90}$ & R & FA$_t$ & $F_{90}$ & R & FA$_t$ \\
\midrule
$H=2$ & $5.5$ & $68.9$ & $0.85$ & $5.5$ & $71.1$ & $0.94$ \\
\rowcolor{oursblue}
$\boldsymbol{H=3}$ & $>5.5$ & $64.4$ & $0.84$ & $5.5$ & $82.2$ & $0.86$ \\
$H=4$ & $>5.5$ & $63.0$ & $0.81$ & $4.5$ & $83.7$ & $1.23$ \\
\bottomrule
\end{tabularx}
\end{minipage}%
\hspace{0.012\textwidth}%
\begin{minipage}[t]{0.309\textwidth}
\centering
\textbf{(d) G1 evaluation window}\par
\vspace{0.2ex}
\setlength{\tabcolsep}{1.1pt}
\begin{tabularx}{\linewidth}{@{}l*{4}{>{\raggedleft\arraybackslash}X}@{}}
\toprule
& \multicolumn{2}{c}{Off} & \multicolumn{2}{c}{On} \\
\cmidrule(lr){2-3}\cmidrule(lr){4-5}
Window & $F_{90}$ & R & $F_{90}$ & R \\
\midrule
$0.10$ s & $>5.5$ & $13.3$ & $>5.5$ & $20.7$ \\
$0.25$ s & $>5.5$ & $38.5$ & $>5.5$ & $57.0$ \\
\rowcolor{oursblue}
\textbf{0.50 s} & $>5.5$ & $64.4$ & $5.5$ & $82.2$ \\
\bottomrule
\end{tabularx}
\end{minipage}
\par
\vspace{0.65ex}
\begingroup
\scriptsize
\setlength{\tabcolsep}{1.1pt}
\begin{minipage}[t]{0.309\textwidth}
\centering
\textbf{(e) Prediction horizon}\par
\vspace{0.2ex}
\begin{tabularx}{\linewidth}{@{}l*{6}{>{\raggedleft\arraybackslash}X}@{}}
\toprule
& \multicolumn{3}{c}{ARX} & \multicolumn{3}{c}{G1} \\
\cmidrule(lr){2-4}\cmidrule(lr){5-7}
Variant & $F_{90}$ & R & FA$_t$ & $F_{90}$ & R & FA$_t$ \\
\midrule
$H=1$ & $>5.5$ & $4.4$ & $0.22$ & $5.5$ & $70.4$ & $0.88$ \\
$H=2$ & $>5.5$ & $46.7$ & $1.26$ & $5.5$ & $71.1$ & $0.94$ \\
\addlinespace[0.25pt]
\rowcolor{oursblue}
$\boldsymbol{H=3}$ & $3.5$ & $74.1$ & $3.32$ & $5.5$ & $82.2$ & $0.86$ \\
$H=4$ & $3.5$ & $74.1$ & $4.55$ & $4.5$ & $83.7$ & $1.23$ \\
$H=5$ & $3.5$ & $80.0$ & $7.92$ & $5.5$ & $81.5$ & $0.95$ \\
\bottomrule
\end{tabularx}
\end{minipage}%
\hspace{0.012\textwidth}%
\begin{minipage}[t]{0.329\textwidth}
\centering
\textbf{(f) Predictor family}\par
\vspace{0.2ex}
\begin{tabularx}{\linewidth}{@{}l*{6}{>{\raggedleft\arraybackslash}X}@{}}
\toprule
& \multicolumn{3}{c}{ARX} & \multicolumn{3}{c}{G1} \\
\cmidrule(lr){2-4}\cmidrule(lr){5-7}
Variant & $F_{90}$ & R & FA$_t$ & $F_{90}$ & R & FA$_t$ \\
\midrule
Persistence & $>5.5$ & $0.0$ & $0.02$ & $>5.5$ & $24.4$ & $0.59$ \\
Ridge & $>5.5$ & $31.1$ & $0.23$ & $>5.5$ & $48.9$ & $0.91$ \\
\addlinespace[0.25pt]
MLP & $3.5$ & $77.8$ & $14.18$ & $4.5$ & $81.5$ & $0.84$ \\
GRU & $5.5$ & $57.0$ & $0.72$ & $2.5$ & $92.6$ & $0.45$ \\
\rowcolor{oursblue}
\textbf{Transformer} & $3.5$ & $74.1$ & $3.32$ & $5.5$ & $82.2$ & $0.86$ \\
\bottomrule
\end{tabularx}
\end{minipage}%
\hspace{0.012\textwidth}%
\begin{minipage}[t]{\dimexpr\textwidth-0.309\textwidth-0.329\textwidth-2\dimexpr0.012\textwidth\relax\relax}
\centering
\textbf{(g) Prediction and evidence}\par
\vspace{0.2ex}
\begin{tabularx}{\linewidth}{@{}l*{6}{>{\raggedleft\arraybackslash}X}@{}}
\toprule
& \multicolumn{3}{c}{ARX} & \multicolumn{3}{c}{G1} \\
\cmidrule(lr){2-4}\cmidrule(lr){5-7}
Variant & $F_{90}$ & R & FA$_t$ & $F_{90}$ & R & FA$_t$ \\
\midrule
\rowcolor{oursblue}
\textbf{Direct} & $3.5$ & $74.1$ & $3.32$ & $5.5$ & $82.2$ & $0.86$ \\
Autoregressive & $4.5$ & $69.6$ & $6.30$ & $5.5$ & $65.9$ & $0.81$ \\
\specialrule{0.25pt}{0pt}{0pt}
\rowcolor{oursblue}
\textbf{Instantaneous} & $3.5$ & $74.1$ & $3.32$ & $5.5$ & $82.2$ & $0.86$ \\
Fast transient & $>5.5$ & $40.0$ & $1.00$ & $>5.5$ & $54.8$ & $0.99$ \\
Slow sustained & $3.5$ & $80.0$ & $26.19$ & $5.5$ & $50.4$ & $0.56$ \\
\bottomrule
\end{tabularx}
\end{minipage}
\endgroup
\caption{\textbf{Ablations of prediction, scoring, and temporal design.} Panels compare (a) state exclusion and residual aggregation, (b) noise weighting, (c) state exclusion across horizons $H$, (d) G1 detection windows with fixed thresholds, (e) prediction horizons, (f) predictor families, and (g) prediction and temporal filtering. Off/On disables/enables state exclusion within the prediction interval. Std. norm denotes the standardized torque residual norm; W./U. Jac. denotes weighted/unweighted Jacobian scoring. $\gamma_T$ and $\gamma_{\rm br}$ are thresholds for task-matched and broad contact-free motion, with $\gamma_B=\max(\gamma_T,\gamma_{\rm br})$ for combined calibration, used in (b)--(g). For G1, this maximum is computed per fold and seed; reported ratios divide the mean broad threshold by the mean task threshold. $F_{90}$ (N) is the lowest tested reference sliding resistance with at least 90\% detection at that and all higher levels; $>5.5$ indicates that the criterion is unmet. R is overall trial recall (\%); FA$_t$ is the percentage of held-out broad contact-free motion spent in confirmed alarm. Blue marks defaults; Direct and Instantaneous share the same configuration. State exclusion and weighted aggregation jointly retain sensitivity under broader calibration.}
\label{fig:ablation}
\end{figure*}

\subsection{Detection across Reference Sliding Resistances (Q1)}

\looseness=-1 Figure~\ref{fig:calibration_force_levels} compares detector scores at $0.04$~m/s, each divided by its calibration threshold so that the decision threshold is 1. Curves show the mean and shading the 10--90\% range across three trials and, for learned methods, three seeds. The vertical lines mark approximate contact references for visual guidance, not measured contact onset. Blue (B) and orange (T) indicate combined and task-matched calibration. Detection rates combine all three speeds; curves show the middle speed.

\vspace{1mm}
\noindent\textbf{ARX detection performance.}
CoPRE retains sensitivity under combined calibration (Fig.~\ref{fig:calibration_force_levels}). It reaches $F_{90}=3.5$~N and 74.1\% recall (95\% CI: 62.2--85.2\%). Both baselines miss all contacts, with $F_{90}>5.5$~N. CoPRE has no confirmed alarms on nine task-matched calibration pushes. On held-out broad contact-free motion, it reduces FA$_t$ from 14.51\% for the neural baseline to 3.32\%. Dynamics yields zero FA$_t$.

\vspace{1mm}
\noindent\textbf{Effect of calibration coverage.}
Combined calibration affects the neural baseline much more than CoPRE. With task-matched calibration, neural detection rises from 22.2\% at 1.5~N to 100\% at 5.5~N, while CoPRE reaches 100\% at 3.5~N. Adding broad contact-free motion raises the neural threshold to 2.48--2.63 times its task-matched value across seeds and eliminates its contact detections. Its normalized curves fall below the threshold, while CoPRE's blue and orange curves remain close (Fig.~\ref{fig:calibration_force_levels}). CoPRE's threshold changes by $\leq7.4\%$; task-matched and combined calibration yield 74.8\% and 74.1\% recall, respectively.

\vspace{1mm}
\noindent\textbf{G1 detection performance.}
CoPRE improves weak-contact sensitivity on the G1 right arm (Fig.~\ref{fig:g1_results}). It reaches $F_{90}=5.5$~N and 82.2\% recall (95\% CI: 72.6--91.1\%), versus 16.3\% for the neural and 42.2\% for the dynamics baseline. Neither baseline attains $F_{90}$ within the tested range. CoPRE detects 55.6\% of trial--seed evaluations at 1.5~N and 85.2\% at 2.5~N. Held-out task/broad FA$_t$ is 1.37/0.86\% for CoPRE and 0.80/0.36\% for the dynamics baseline.

\begin{figure*}[t]
    \centering
    \includegraphics[width=\textwidth,trim=0 0.25cm 0 0,clip]{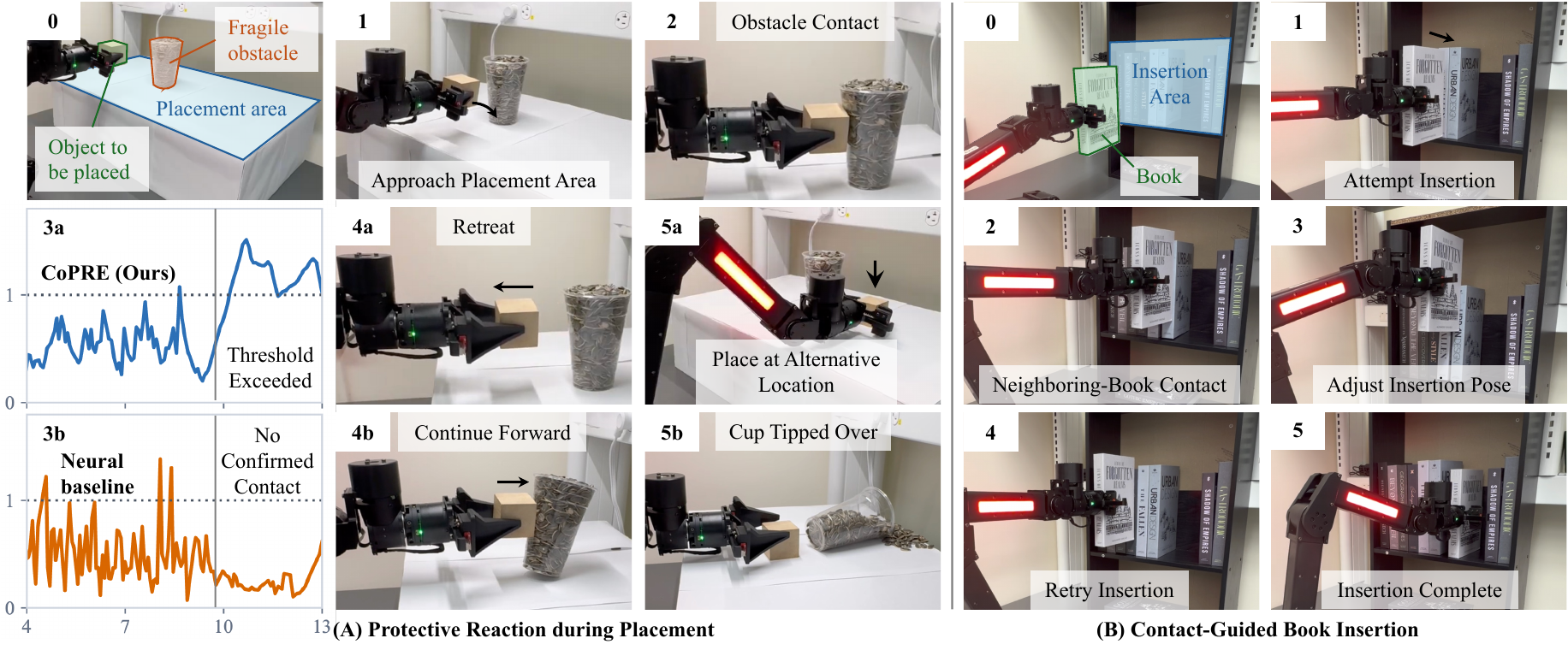}
    \caption{\textbf{Contact-guided manipulation on ARX.} (A) A cup obstructs placement of a carried block. CoPRE triggers retreat and alternate placement, leaving the cup upright (3a--5a); the neural baseline continues forward and tips it over (3b--5b). Insets: threshold-normalized detector scores; dotted lines: threshold 1. (B) Contact with a neighboring book prompts pose adjustment and a successful insertion retry. Both sequences are qualitative demonstrations.}
    \label{fig:downstream_tasks}
\end{figure*}

\subsection{Ablation Study (Q2)}

Figure~\ref{fig:ablation} tests state exclusion, residual aggregation, prediction design, and evaluation-window duration. Each comparison holds the data split, reference sliding resistances, confirmation rule, and calibration protocol fixed except for the named factor.

\vspace{1mm}
\noindent\textbf{State exclusion and residual aggregation.}
\looseness=-1 The $2\times2$ comparison varies access to state observations within the prediction interval and the residual score: the weighted Jacobian score from Eq.~\eqref{eq:force_proxy}, or the standardized torque residual norm $\lVert D_\sigma^{-1}r_t\rVert_2$. Within each robot, the variants share the transformer architecture, $H=3$, training data, loss, and residual-normalization procedure. With state exclusion off, the predictor receives the preceding measured state within the prediction interval; with it on, measured states end at the earlier anchor. Both robots use task-matched $\gamma_T$ and combined $\gamma_B=\max(\gamma_T,\gamma_{\rm br})$ thresholds, with $\gamma_{\rm br}$ calibrated on broad contact-free motion.

\looseness=-1 On ARX, state exclusion increases task-matched recall by 34.8 percentage points with the weighted Jacobian score (95\% CI: 22.2--46.7) and 32.6 with the standardized residual norm (95\% CI: 20.0--45.9). Under combined calibration, the standardized norm needs a much higher threshold and loses all detections. With state exclusion, the weighted Jacobian score retains sensitivity with little threshold change, supporting the complementary roles of state exclusion and aggregation.

\looseness=-1 On G1, state exclusion improves recall with either score under both calibration scopes. With state exclusion, combined calibration lowers standardized-norm recall from 49.6\% to 31.9\%, reducing broad FA$_t$ from 5.40\% to 1.02\%. Weighted Jacobian recall changes only from 83.0\% to 82.2\%, with broad FA$_t$ of 0.92\% and 0.86\%. Together, these results show how state exclusion provides sensitivity and weighted aggregation preserves it under broader calibration. The state-exclusion gain also persists across $H=2$--$4$ (Fig.~\ref{fig:ablation}(c)); at $H=1$, both variants use identical inputs.

\vspace{1mm}
\noindent\textbf{Aggregation and noise weighting.}
Figure~\ref{fig:ablation}(b) separates geometric mapping from noise weighting by holding predictors, kinematics, residual centering, relative damping, and calibration rules fixed. The unweighted Jacobian score sets all residual scales to one. Noise weighting improves sensitivity on both robots while increasing broad FA$_t$.

\vspace{1mm}
\noindent\textbf{Evaluation-window sensitivity.}
With predictors and thresholds fixed, shortening the G1 evaluation window from 0.50 to 0.10~s lowers CoPRE recall from 82.2\% to 20.7\%. State exclusion improves recall at every tested duration (Fig.~\ref{fig:ablation}(d)).

\vspace{1mm}
\noindent\textbf{Prediction design.}
\looseness=-1 The horizon sweep shows the trade-off around the common setting $H=3$ (Fig.~\ref{fig:ablation}(e)). Shorter horizons reduce recall on both robots. On G1, $H=4$ lowers $F_{90}$ from 5.5 to 4.5~N but raises broad FA$_t$ from 0.86\% to 1.23\%. Longer horizons retain ARX $F_{90}$ and raise FA$_t$.

\looseness=-1 Predictor families share 20 history samples, $H=3$ command input, full-state targets, aggregation, and the decision rule (Fig.~\ref{fig:ablation}(f)). The transformer matches the MLP's ARX $F_{90}$ with much lower FA$_t$, while GRU improves both G1 sensitivity and FA$_t$. We use the transformer on both robots to keep the default architecture consistent.

\flushcolsend
Direct multi-step prediction improves recall relative to autoregressive rollout on both robots (Fig.~\ref{fig:ablation}(g)). Temporal filtering has a less consistent benefit. The fast variant subtracts an exponentially smoothed reference (0.35\,s), smooths the squared deviation over 0.1\,s, and takes its square root. The slow variant smooths the vector over 0.3\,s before taking its norm. Both use their 99th percentile on contact-free training data for normalization. Fast evidence loses sensitivity on both robots; slow evidence increases ARX recall with a large increase in FA$_t$ and lowers G1 FA$_t$ at a sensitivity cost. These comparisons support direct multi-step prediction and instantaneous scoring as the benchmark default.

\vspace{1mm}
\noindent\textbf{Calibration trade-off.}
A retrospective sweep with frozen predictors illustrates the operating trade-off. On ARX, relaxing the calibration FA$_t$ budget from 0\% to 1\% raises recall from 74.1\% to 95.6\%, while held-out broad FA$_t$ rises from 3.32\% to 12.41\%. On G1, tightening the budget from 1\% to 0.1\% lowers task/broad FA$_t$ from 1.37/0.86\% to 0.26/0.14\%, with recall decreasing from 82.2\% to 58.5\%. The calibration budget should balance missed contacts and unnecessary task interruptions; held-out FA$_t$ can still exceed this budget.

\subsection{Contact-Guided Manipulation (Q3)}

\vspace{1mm}
\noindent\textbf{Task settings.}
We qualitatively evaluate belief-guided planning on ARX for object placement around a cup and book insertion among neighboring books (Fig.~\ref{fig:downstream_tasks}).

\vspace{1mm}
\noindent\textbf{Belief-guided task policy.}
\looseness=-1 Detected contact stops blocked motion and prompts replanning (Fig.~\ref{fig:method}(C)). After each attempt, the contact decision, robot pose, and task progress update a spatial belief map: contact increases the estimated probability of an obstruction nearby, while traversed path segments provide evidence of free space. After retreat, the planner ranks placement or insertion attempts by estimated feasibility, distance to the goal, and previous failures.

\vspace{1mm}
\noindent\textbf{Protective reaction during placement.}
\looseness=-1 A movable cup blocks the arm as it places a rigid block. CoPRE triggers a stop and retreat, followed by placement at an alternate location, leaving the cup upright (Fig.~\ref{fig:downstream_tasks}(A), 3a--5a). With the neural baseline, the arm continues and tips the cup over (3b--5b).

\vspace{1mm}
\noindent\textbf{Contact-guided book insertion.}
During book insertion, contact with a neighboring book blocks the first attempt. Contact feedback prompts a pose adjustment, and the next attempt succeeds (Fig.~\ref{fig:downstream_tasks}(B)). Task-success rates remain unquantified.

\section{Discussion and Conclusion}
\label{sec:conclusion}

\looseness=-1 CoPRE improves weak-contact sensitivity through state exclusion and noise-weighted Jacobian aggregation. It achieves $F_{90}=3.5$~N on ARX and $5.5$~N on G1; neither baseline meets this criterion under the same calibration rule within each robot. Ablations support both designs. Qualitative ARX demonstrations show how contact feedback supports protective retreat during object placement and pose adjustment for book-insertion retries.

\looseness=-1 CoPRE requires robot-specific contact-free training and calibration. Sensitivity to data quantity, motion diversity, unseen motions, and payload changes remains untested; false-alarm rates may change in new conditions. During sustained contact, affected states can enter the input history, allowing the reference to follow torque changes and reduce the residual. Evaluation covers contacts transmitted through the end effector. Future work will test other arm links and quantify detection latency, task success, and false interruptions.

\clearpage
\raggedcolsend
\raggedend
\bibliographystyle{IEEEtran}
\bibliography{references}

\end{document}